\documentclass[final,runningheads]{llncs}
\usepackage[T1]{fontenc}
\usepackage{graphicx}

\usepackage{amsfonts} \usepackage{amsmath} 
\usepackage{amssymb} \usepackage{bbm} \usepackage{tikz}
\usetikzlibrary{shapes.geometric, arrows.meta, positioning}

\tikzstyle{box} = [rectangle, rounded corners, minimum width=3cm, minimum height=1cm, text centered, draw=black]
\tikzstyle{arrow} = [thick,->,>=stealth]

\usepackage{longtable}
\usepackage{booktabs} 

\usepackage{graphicx}
\usepackage{subcaption}
\usepackage{verbatim}
\usepackage{multirow}
\usepackage[misc]{ifsym}
\usepackage[sort]{cite}

\usepackage{algorithm}
\usepackage{algorithmicx}
\usepackage[noend]{algpseudocode}

\algnewcommand{\LineComment}[1]{\Statex $\triangleright$ #1}
\algnewcommand\algorithmicinput{\textbf{Input:}}
\algnewcommand\algorithmicoutput{\textbf{Output:}}
\algnewcommand\Input{\item[\algorithmicinput]}
\algnewcommand\Output{\item[\algorithmicoutput]}

\def\algbackskip{\hskip-\ALG@thistlm}

\usepackage[hidelinks]{hyperref}
\usepackage{orcidlink}

\newcommand{\cmmnt}[1]{}

\newcommand{\itadata}{\footnotesize \textsl{ITADATA2024: The 3$^{\text{rd}}$ Italian Conference on Big Data and Data Science}}
\usepackage{fancyhdr}
\fancypagestyle{empty}{\fancyhf{}\fancyhead[C]{\itadata}
  \fancyhead[R]{\includegraphics[width=.05\textwidth]{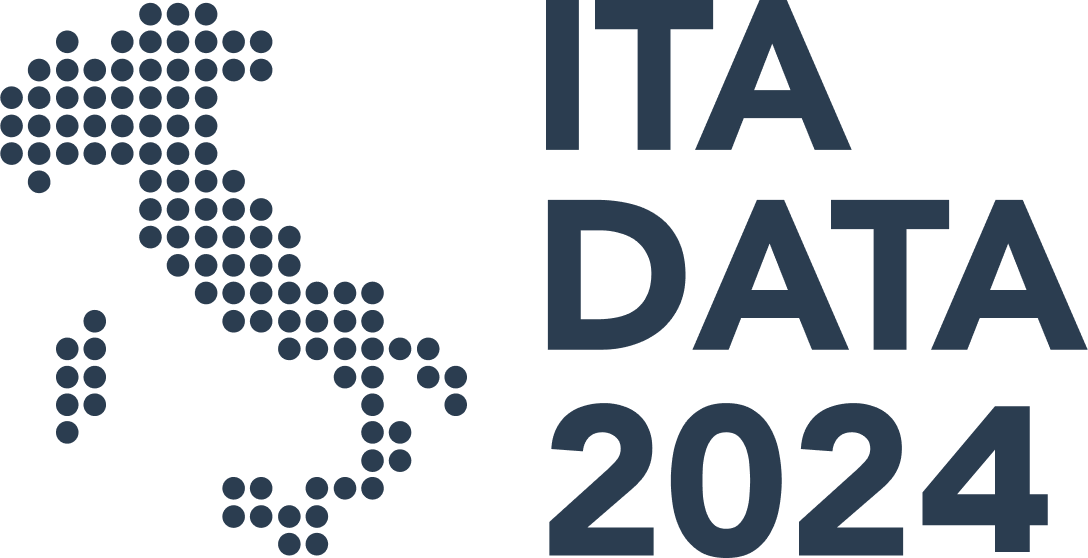}}}

\makeatletter
\begin{document}

\setlength{\parskip}{0pt}

\title{Interpretable Machine Learning\\ for Oral Lesion Diagnosis through\\ Prototypical Instances Identification}

\author{Alessio Cascione\inst{1}\orcidlink{0009-0003-5043-5942} \and
    Mattia Setzu\inst{1}\orcidlink{0000-0001-8351-9999} \and
    Federico A. Galatolo\inst{1}\orcidlink{0000-0001-7193-3754} \and \\
    Mario G.C.A. Cimino \inst{1}\orcidlink{0000-0002-1031-1959} \and
    Riccardo Guidotti\inst{1,2}\orcidlink{0000-0002-2827-7613}
}

\authorrunning{Alessio Cascione et al.}

\institute{
    University of Pisa, Largo Bruno Pontecorvo 3, Pisa PI 56127, Italy \email{a.cascione@studenti.unipi.it},\email{\{name.surname\}@unipi.it} \and
    KDD Lab, ISTI-CNR, Via G. Moruzzi 1, Pisa PI 56124, Italy \email{riccardo.guidotti@isti.cnr.it}
}

\maketitle              \begin{abstract}
Decision-making processes in healthcare can be highly complex and challenging. 
Machine Learning tools offer significant potential to assist in these processes. 
However, many current methodologies rely on complex models that are not easily interpretable by experts. 
This underscores the need to develop interpretable models that can provide meaningful support in clinical decision-making. 
When approaching such tasks, humans typically compare the situation at hand to a few key examples and representative cases imprinted in their memory. 
Using an approach which selects such exemplary cases and grounds its predictions on them could contribute to obtaining high-performing interpretable solutions to such problems. 
To this end, we evaluate \textsc{PivotTree}, an interpretable prototype selection model, on an oral lesion detection problem, specifically trying to detect the presence of \textit{neoplastic}, \textit{aphthous} and \textit{traumatic} ulcerated lesions from oral cavity images. 
We demonstrate the efficacy of using such method in terms of performance and offer a qualitative and quantitative comparison between exemplary cases and ground-truth prototypes selected by experts.

\end{abstract}

\section{Introduction}
One of the sectors that has significantly benefited from the application of Machine Learning (ML) tools is healthcare~\cite{Celard2022ex2,Javaid2022sign}.
However, although the models employed to solve diagnostic tasks are powerful in terms of predictive capability, their reliance on complex architectures often makes it difficult for experts and users to understand their reasoning. 
Moreover, the ``cognitive process'' employed by these models is frequently not comparable to how humans reason to solve the same tasks~\cite{yang2022unbox}. 
Given the pivotal role of these tools as decision-support systems for practitioners in healthcare, explaining and interpreting their predictions has become crucial and is the focus of active research in Explainable AI (XAI)~\cite{enlig2023}.

As humans, our cognitive processes and mental models frequently depend on case-based reasoning~\cite{schank2014knowledge}, where past exemplary cases are stored in memory and retrieved to solve specific tasks. 
Especially in healthcare, practitioners often perform diagnosis or identify new conditions by relying on past case reports~\cite{Harasym2008,Shin2019medicalreas}. 
Given these premises, a promising approach to designing inherently interpretable ML models for the healthcare sector is to explore the intuitive notion of similarity between \textit{discriminative} and \textit{representative} instances. 
The underlying assumption is that grounding a model's predictions on the similarity between test instances and exemplar cases would yield a naturally interpretable and trustworthy tool for medical experts and end-users alike.
In this paper, we present a case study with an interpretable similarity-based model for decision-making applied to a specific medical context, i.e., for an oral lesion prediction task. 

In particular, we study \textsc{PivotTree}~\cite{cascione2024pivotree}, a hierarchical and interpretable case-based model inspired
by Decision Tree (\textsc{DT})~\cite{breiman1984classification}. 
By design, \textsc{PivotTree} can be used both as a \emph{prediction} and \emph{selection} model. 
As a selection model, \textsc{PivotTree} identifies a set of training exemplary cases named \emph{pivots};
as a predictive model, \textsc{PivotTree} leverages the identified pivots to build a similarity-based \textsc{DT}, routing instances through its structure and yielding a prediction, and an associated explanation.
Unlike traditional \textsc{DT}s, the resulting explanation is not a set of rules having features as conditions, but rules using a set of pivots to which the instance to predict is compared. 
Like distance-based models, \textsc{PivotTree} allows to select exemplary instances in order to encode the data in a similarity space that enables case-based reasoning. 
Finally, \textsc{PivotTree} is a \emph{data-agnostic} model, which can be applied to different data modalities, jointly solving both pivot selection and prediction tasks.
Given its modality agnosticism, \textsc{PivotTree} represents an advancement over traditional \textsc{DT}s. 
As shown in~\cite{cascione2024pivotree}, the case-based model learned by \textsc{PivotTree} offers interpretability even in domains like images, text, and time series, where conventional interpretable models often underperform and lack clarity. 
Furthermore, unlike conventional distance-based predictive models such as k-Nearest Neighbors (\textsc{kNN})~\cite{fix1985discriminatory}, \textsc{PivotTree} introduces a hierarchical structure to guide similarity-based predictions.

\begin{figure}[t]
    \centering
    \begin{subfigure}[b]{0.32\textwidth}
        \centering
        \includegraphics[width=\textwidth]{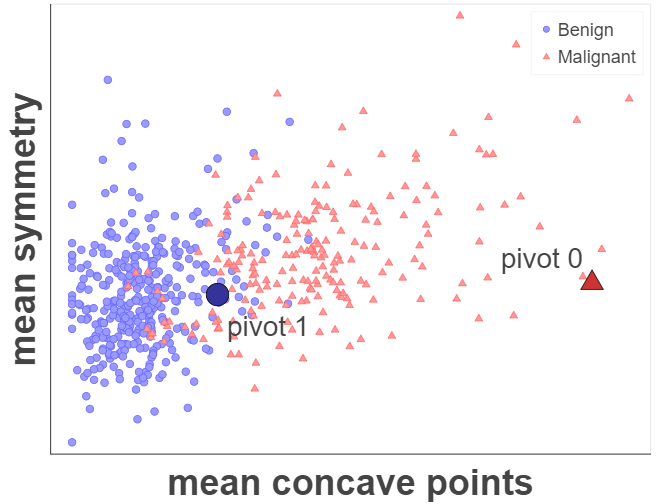}
        \caption{Select}
        \label{fig:pivots}
    \end{subfigure}
    \hfill
    \begin{subfigure}[b]{0.32\textwidth}
        \centering
        \includegraphics[width=\textwidth]{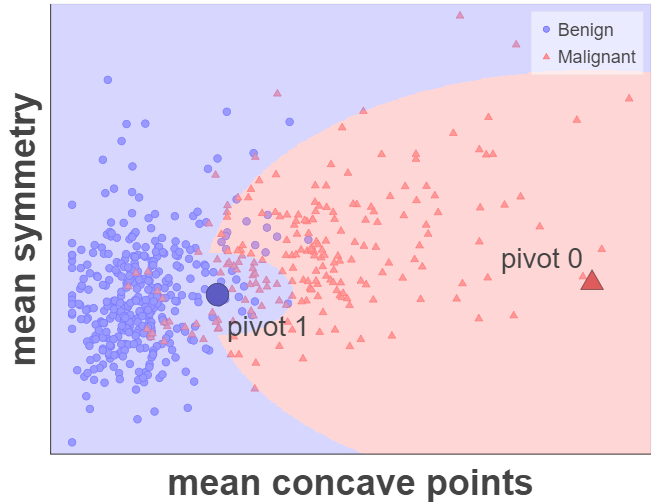}
        \caption{Predict}
        \label{fig:example:similarities}
    \end{subfigure}
    \hfill
    \begin{subfigure}[b]{0.32\textwidth}
        \centering
        \includegraphics[width=\textwidth]{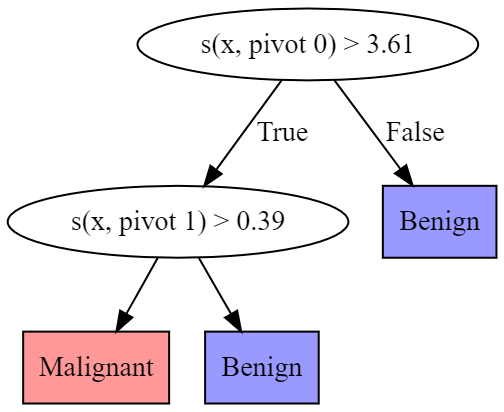}
        \caption{Explain}
        \label{fig:example:tree}
    \end{subfigure}
    \hfill    
    \caption{\textsc{PivotTree} as \textit{(a)} selector, \textit{(b)} interpretable model, \textit{(c)} Decision Tree.}
    \label{fig:example}
\end{figure}

Fig.~\ref{fig:example} provides an example of \textsc{PivotTree} on the \texttt{breast cancer} dataset\footnote{https://archive.ics.uci.edu/dataset/17/breast+cancer+wisconsin+diagnostic}, wherein cell nuclei are classified according to their characteristics computed from a digitized image of a fine needle aspirate of a breast mass. Starting from a dataset of instances, \textsc{PivotTree} identifies a set of two pivots (Fig.~\ref{fig:example} \textit{(a)}) in this case belonging to the two distinct classes \textit{Benign} and \textit{Malignant}.
Said \textit{pivots} are used to learn a case-based model wherein novel instances are represented in terms of their similarity to the induced pivots (Fig.~\ref{fig:example} \textit{(b)}).
Building on pivot selection, \textsc{PivotTree} then learns a hierarchy of pivots wherein instances are classified.
This hierarchy takes the form of a Decision Tree (Fig.~\ref{fig:example} \textit{(c)}): novel instances navigate the tree, gravitating towards pivots to which they are more similar or dissimilar, and landing into a classification leaf.
In the example, given a test instance $x$: if its similarity to \textit{pivot 0} is lower than 3.61 (following the right branch), then $x$ is classified as a \textit{Benign}, i.e., $x$ is far away from the \textit{Malignant} \textit{pivot 0} (see Fig.~\ref{fig:example} \textit{(b)}).
Instead, following the left branch, if $x$'s similarity to  \textit{pivot 1} is higher than 0.39 (left branch), then $x$ is still classified as \textit{Benign} as it is very similar to the \textit{Benign} \textit{pivot 1}, otherwise $x$ is classified as \textit{Malignant} as it is sufficiently similar to the \textit{Malignant} \textit{pivot 0}.
In contrast, a traditional Decision Tree (DT) would model the decision boundary with feature-based rules, e.g., ``if \textit{mean concave points} < 2.4 then \textit{Benign} else if \textit{mean symmetry} < 1.7 then \textit{Malignant}''.
However, traditional DTs \emph{(i)} can only model axis-parallel splits, and \emph{(ii)} cannot be employed on data types with features without clear semantics such as medical images.
Hence, improving on traditional DTs, the case-based model learned by \textsc{PivotTree} can provide interpretability even in domains such as images, text, and time series, by exploiting a suitable data transformation. 

In this paper we demonstrate that \textsc{PivotTree} represents an effective approach for \textit{interpretability of oral lesion detection}, and we compare its selected pivots with instances identified as representative by domain experts. After a review of the literature concerning XAI in the healthcare sector, and prototype-based approach for explainability in Section~\ref{sec:related}, in Section~\ref{sec:method} we summarize the \textsc{PivotTree} method. Then, in Section~\ref{sec:experiments} we report the experimental results on the oral lesion diagnostic problem. Finally, Section~\ref{sec:conclusion} completes our contribution and discusses future research directions.

\section{Related Work}
\label{sec:related}
The wide use of explainability techniques for the medical field has been extensively reviewed in previous work~\cite{Frasca2024,SBand2023exp2}.
ML~\cite{Dixit2023ex}, and specifically case-based reasoning, already finds  application in the medical domain, where interpretable and uninterpretable models~\cite{healthcase2006,Choudhury2016} already tackle a variety of tasks, including
breast cancer prediction~\cite{Lamy2019},
oral cancer detection~\cite{kouketsu2024detection,welikala2020automated,zhou2024pathology},
melanoma detection~\cite{exemplars2021metta,metta2024advancing,metta2023improving},
and Covid-19 detection~\cite{Singh2021genprotonet}.
Case-based models, which leverage similarity to a set of prototypes, may vary in how such prototypes and similarity are defined, and in the heterogeneity of the prototypes themselves, some models focusing on improving similarity computation~\cite{chen2019protopnet,Singh2021genprotonet}, others focusing on increasing heterogeneity of prototypes~\cite{Kim2021XProtoNetDI}
The latter, in particular, introduces two-level interpretations: prototypes are also defined contrastively, i.e., both highly similar and highly dissimilar prototypes are provided, and they are also accompanied by heatmaps indicating regions of higher importance.
These approaches integrate the discovery of prototypes directly into the model, which often uses similarity-based scoring function to perform predictions.
In~\cite{kim2016examples} besides prototypes criticism are also identified, i.e., instances representatives of some parts of the input space where prototypical examples do not provide good explanations.

A case-based approach specifically for oral lesion is offered in~\cite{ehtesham2019casebased}, which works with tabular descriptors by physicians. 
More at large, and aside from case-based interpretations, interpretability in the medical sector has been gaining attention for quite some years~\cite{DBLP:conf/fat/PaniguttiPP20}.
In terms of interpretability tools for oral cancer detection, only a handful of proposals are currently in place.
In~\cite{Figueroa2022} an approach using gradient-weighted class activation mapping is presented and~\cite{Song2023ex2} provides visual explanations leveraging attention mechanisms.
To our knowledge, our study is the first inquiring on explainability through prototypes for the oral lesion detection problem using a data-agnostic model.

\section{Pivot Tree in a Nutshell}
\label{sec:method}
We present the main characteristics of \textsc{PivotTree}: for more detailed information and benchmarking, we refer readers to~\cite{cascione2024pivotree}.
Given a set of $n$ instances represented as real-valued $m$-dimensional feature vectors\footnote{For the sake of simplicity, we consistently treat data instances as real-valued vectors. Any data transformation employed in the experimental section to maintain coherence with this assumption will be specified when needed.}
in $\mathbb{R}^{m}$, and a set of class labels $C = \{1, \ldots, c\}$, in case-based reasoning, the objective is to learn a function $f: \mathbb{R}^m \rightarrow C$ approximating the underlying classification function, with $f$ being defined as a function of $k$ exemplary cases named \textit{pivots}.
Similarity-based case-based models define $f$ on a similarity space $\mathcal{S}$, often inversely denoted as ``distance space'', induced by a similarity function $s: \mathbb{R}^m \times \mathbb{R}^m \rightarrow \mathbb{R}$ quantifying the similarity of instances~\cite{pekalska2005diss}.
Given a training set $\langle X, Y \rangle$, and a similarity function $s$, our objective is to learn a function $\pi: \mathbb{R}^{n\times m} \rightarrow \mathbb{R}^{k \times m}$ that selects a set $P \subseteq X$ of $k$ pivots maximizing the performance of $f$.
The instances in $X$ are mapped into a similarity-based representation through $\mathcal{S}$, wherein they are represented in terms of their similarity to the pivots $P$. 

This similarity-based dataset $Z \in \mathbb{R}^{\mid X \mid \times \mid P \mid}$ holds in $Z_{i,j}$ the similarity between the $i$-th instance in $X$ and the $j$-th pivot in $P$. The predictive model $f$ is then trained on $\langle Z, Y \rangle$. To perform inference on a test instance $x \in \mathbb{R}^m$, $x$ is first mapped to a similarity vector $z = \langle s(x, p_1), \dots, s(x, p_k) \rangle$ yielding its similarity to the set $P$ of pivots; then, $z$ is provided to $f$, which performs the prediction.
Aiming for transparency of the case-based predictive model $f$, our objective is to employ as an interpretable model $f$ Decision Tree classifiers (\textsc{DT}) or k-Nearest Neighbors approaches~\cite{guidotti2019survey} (kNN).
When $f$ is implemented with a \textsc{DT}, split conditions will be of the form $s(x, p_i) \geq \beta$, i.e., ``if the similarity between instance $x$ and pivot $p_i$ is greater or equal then $\beta$, then ...'', allowing to easily understand the logic condition by inspecting $x$ and $p_i$ for every condition in the rule.

On the other hand, when $f$ is implemented as a kNN, every decision will be based on the similarity with a few neighbors derived from the pivot set $P$.
A human user just needs to inspect $x$ and the similarities with the pivots $P$ and the instances in the neighborhood. 
When the number of pivots is kept small, the interpretability of both methods increases, limiting the expressiveness.
Vice versa, using a selection model $\pi$ that returns a large number $k$ of pivots can increase the performance at the cost of interpretability. \textsc{PivotTree} implements the selection function $\pi$, and leverages existing interpretable models to implement $f$.

Much like Decision Tree induction algorithms~\cite{breiman1984classification}, \textsc{PivotTree} greedily learns a hierarchy of nodes wherein pivots lie.
Node splits are selected so that the downstream performance of $f$ is maximized, i.e., the split is chosen to maximize the information gain of the node.
The training data is then routed according to the split, and the operation repeats recursively. More specifically, during training, each node describes a subset of training instances defined by the decision path leading to that node at a specific iteration.
For these instances, a set of candidate pivots is selected.
The similarity-based split that results in the maximum information gain among the candidates is then used to split the node, routing the instances accordingly to the child nodes. 

Among candidates, we distinguish between \textit{discriminative} pivots, which guide instances through the tree, and \textit{representative} pivots, which instead describe the node.
The former are selected to maximize the performance, while the latter are selected to maximize similarity to the other instances traversing the node.
In a sense, the \textit{representative} and \textit{discriminative} pivots extracted by \textsc{PivotTree} can be associated with the prototypical examples and criticisms identified by~\cite{kim2016examples}.  However, their usage is markedly different.

\textit{Representative} pivots for each class are selected as the instances described by a node that have the highest similarity with all other instances described by the same node and within the same class. Conversely, \textit{discriminative} pivots are chosen to be the instances from each class which best separate the training data described by a node when instance similarity is taken into account, i.e., when the optimal splitting feature is chosen w.r.t. the induced similarity space. Both types of pivots form the candidate set used to determine the actual split of the current training set.
{The process naturally results in a structure of decision rules that can be directly used as a classification model for prediction. At the same time, it selects pivots from increasingly fine-grained partitions of the training data, which can be employed by other transparent models implementing $f$.

By design, \textsc{PivotTree} is a data-agnostic model that leverages the concept of similarity to conduct both selection and prediction tasks simultaneously.
While some data types, e.g., relational data, are more amenable than others, e.g., images or text, to similarity computation, with our contribution, we aim to address all data types as one.
By decoupling similarity computation and object representation, \textsc{PivotTree} can be applied to any data type supporting a mapping to $\mathbb{R}^m$, i.e., text through language model embedding, images through vision models, graphs through graph representation models, etc.
In the following experimentation, we focus exactly on images and on particular on oral lesion images though an embedding provided by a pre-trained deep learning model.

\section{Experiments}
\label{sec:experiments}
In this section, we evaluate the performance of \textsc{PivotTree}\footnote{An implementation regarding the experiments described in Sec.~\ref{sec:experiments} on the oral lesion detection task is available at \url{https:/github.com/acascione/PivotTree_DoctOral}}
(\textsc{PTC}) on the DoctOral-AI dataset\footnote{\url{https://mlpi.ing.unipi.it/doctoralai/}}. 
Our objective is to demonstrate that \textsc{PivotTree} is an accurate predictor and selector tool for the task and show how comparable the learned pivots are to ground-truth cases deemed prototypical by expert doctors. 

\textbf{Classification Models.} We refer to \textsc{PivotTree} used as \textsc{C}lassification model with \textsc{PTC}. 
We use $P$ to denote the set of pivots identified by \textsc{PivotTree}, and $O$ to denote the set of ground-truth prototypes.
$\textsc{DT}_P$ and $\textsc{kNN}_P$ refer to \textsc{DT} and \textsc{kNN} models, respectively, trained in the similarity space obtained by computing the similarity between each instance and every pivot in $P$.
Similarly, $\textsc{DT}_O$ and $\textsc{kNN}_O$ are trained in the similarity space derived from the ground-truth prototypes in $O$.
As further baselines, we compare \textsc{PivotTree} with \textsc{kNN} and \textsc{DT} directly trained on the original feature space.
Finally, as deep learning model we rely on the Detectron2 (\textsc{D2}) model~\cite{wu2019detectron2} fine-tuned on the DoctOral-AI dataset.
We report the performance of \textsc{D2} to observe the loss in accuracy at the cost of interpretability.

\textbf{Experimental Setting.} We evaluated the predictive performance of the aforementioned models by measuring Balanced Accuracy and F1-score, Precision and Recall by computing the metric for each label and reporting the unweighted mean. 
In line with~\cite{cascione2024pivotree}, for \textsc{PivotTree} hyperparameter selection\footnote{For every tree, we set 3 as min nbr. of instances a node must have to be considered leaf, and 5 as the min nbr. of instances a node must have to perform a split.}, both as a predictor and a selector, we aim to maintain a low number of pivots and an interpretable classifier structure. 
Empirical studies~\cite{Huysmans2011dtunderstand} have shown that, for binary classification tasks, using \textsc{DT}s with more than 16 leaves—and therefore depths greater than 4—leads to significant decreases in human subjects' accuracy and confidence when answering logical YES-NO questions about the model's decision structure. Additionally, response times are notably longer with such deeper trees. Therefore, to ensure interpretability, the optimal $\mathit{maxdepth}$ is searched within the interval $[2, 4]$. We focus solely on depth and the number of pivots as measures of interpretability, as explanations taking into account features sparsity, such as explanation size~\cite{Souza2022expsize}, are not directly applicable to \textsc{PivotTree}, due to the \textit{case-based} nature of rules. We plan to extend this approach and develop specific interpretability metrics for \textsc{PivotTree} in future works.

When using \textsc{PivotTree} as a selector, we assess the performance of using different pivot types -- \textit{discriminative}, \textit{representative}, both, and using only those considered as splitting pivots -- to identify which combination achieves the best selection performance when paired with \textsc{DT} or \textsc{kNN}. 
The best performance for $\textsc{kNN}_P$ are obtained with $\mathit{maxdepth} = 3$, while for \textsc{PTC} and $\textsc{DT}_P$ with $\mathit{maxdepth} = 4$.
Leveraging both discriminative and representative pivots consistently yields better results.
Finally, for the baseline \textsc{DT} and \textsc{kNN} the best performance is achieved with $\mathit{maxdepth} = 4$ and $k = 5$, respectively, both in the original space and in the similarity feature space.
As distance function, we always adopt the Euclidean distance.

\textbf{Dataset and Embedding Model.}
The DoctOral-AI dataset comprises 535 images of varying sizes,
which define a multiclassification oral lesion detection task with classes
\textit{neoplastic} (31.58\%), \textit{aphthous} (32.52\%), and \textit{traumatic} (35.88\%).
The dataset is divided into 70\% development and 30\% testing, the former further divided on a 80\%/20\% split for training and validation.
We embed images with a Detectron2 (\textsc{D2})~\cite{wu2019detectron2} CNN architecture fine-tuned on the DoctOral-AI\footnote{We offer details regarding the training process in \url{https://github.com/galatolofederico/oral-lesions-detection}}.
We resized each image into an 800x800 format.
Then relevant feature maps are selected from the \textsc{D2}'s backbone output and passed to the \textsc{D2}'s region of interest pooling layer. 
Finally, a pooling layer and a flattening layer map the feature maps to a 256-dimensional embedding.
We also report the performance of \textsc{D2} to observe the loss in accuracy at the cost of interpretability.

\begin{figure}[t]
    \centering
    \includegraphics[width=1\linewidth]{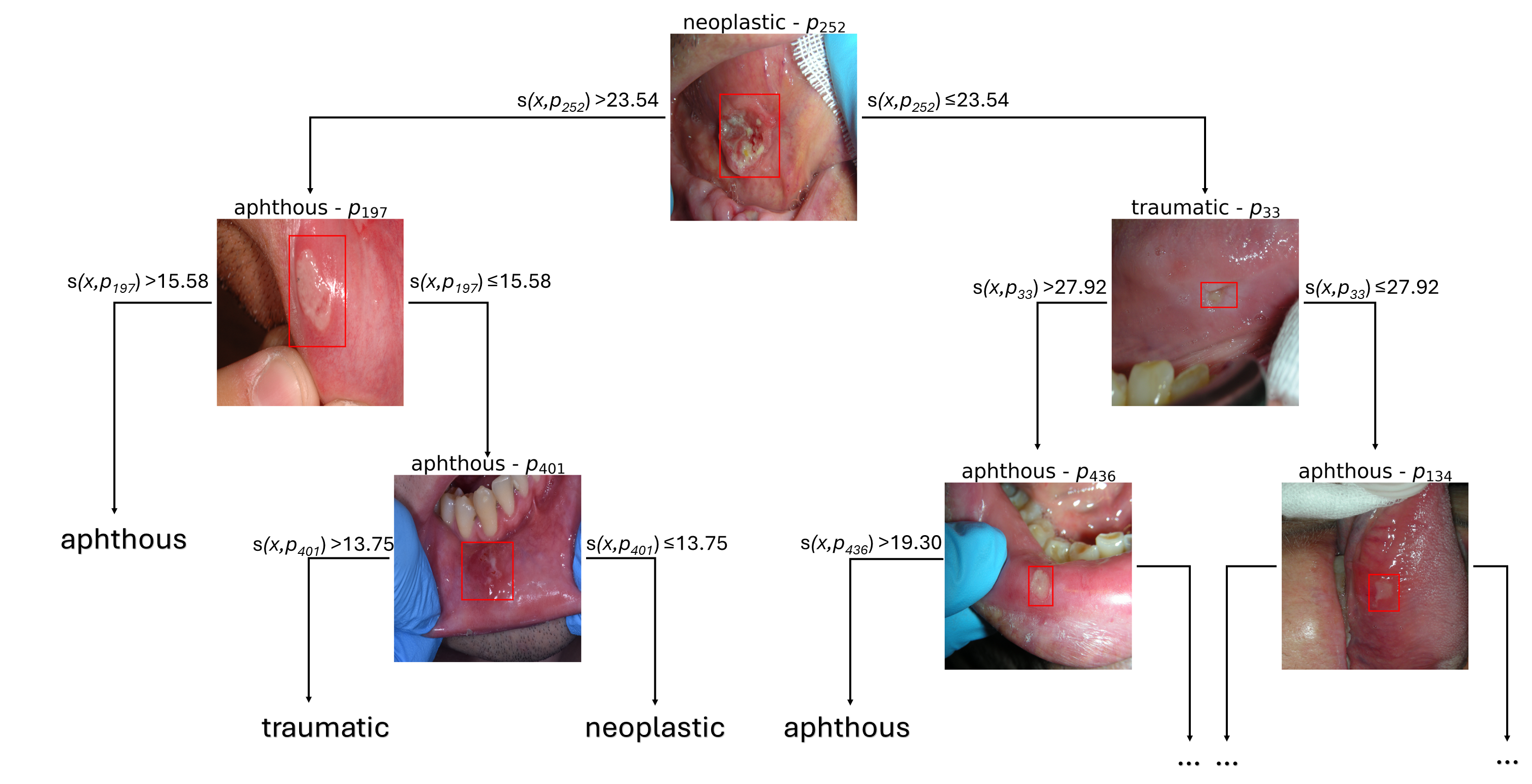}
    \caption{Partial visual depiction of best \textsc{PTC} configuration on the test set. Branches are labeled with similarity threshold values used for prediction.}
    \label{fig:pivotree_example}
\end{figure}

\textbf{Qualitative Results.}
Fig.~\ref{fig:pivotree_example} depicts a visual representation of \textsc{PTC} decision rules and splitting pivots associated with the initial nodes\footnote{The actual trained tree has a $\mathit{maxdepth}$ of 4. For visualization purposes, we limit the visualization to the initial nodes.}. 
Given a hypothetical instance $x$ to predict, the predictive reasoning employed by the trained model proceeds as follows: $x$ is first compared to $p_{252}$,
a \textit{neoplastic} instance. 
If the similarity between $x$ and $p_{252}$ is sufficiently high, then $x$ traverses the left branch and is compared to the \textit{aphthous} pivot $p_{197}$. 
If $x$ is sufficiently similar to $p_{197}$, the model concludes the prediction and assigns $x$ to the \textit{aphthous} class. 
Otherwise, an additional comparison with $p_{401}$ is performed, leading to a final classification as either \textit{neoplastic} or \textit{traumatic}. 
We underline that the path leading to \textit{traumatic} decision lacks pivots belonging to such class. 
This suggests that the model can effectively perform comparisons with pivots belonging to other classes to exclude their possibility for $x$, thereby assigning $x$ to the remaining class by exclusion\footnote{We intend to fix this (possible) issue by extending \textsc{PivotTree} with Proximity Trees~\cite{Lucas2019proximityfor} to compare the test $x$ against two pivots instead of only one.}.
On the other hand, if the initial comparison identifies $x$ as dissimilar from the \textit{neoplastic} $p_{252}$, the model then compares it to the \textit{aphthous} $p_{33}$ and applies analogous reasoning for subsequent comparisons.

\begin{table}[t]
    \centering
    \setlength{\tabcolsep}{3.5mm}
    \caption{Mean predictive performance and number of pivots. Best performer in \textbf{bold}, second best performer in \textit{italic}, third best performed \underline{underlined}.}
    \begin{tabular}{ccrrrrc}
    \toprule
     & Model & Bal. Acc. & F1-score & Precision & Recall & Nbr. Pivots \\
    \midrule
     & \textsc{D2} & \textbf{0.859} & \textbf{0.854} & \textbf{0.854} & \textbf{0.858} & - \\
    \midrule
     & \textsc{PTC} & \textit{0.834} & \textit{0.832} & \textit{0.839} & \textit{0.834} & \textit{9} \\
     & $\textsc{DT}_{P}$ & \underline{0.833} & \underline{0.830} & \underline{0.830} & \underline{0.833} & \underline{47} \\
     & $\textsc{kNN}_{P}$ & 0.811 & 0.807 & 0.810 & 0.811 & \textbf{5} \\
    \midrule
     & $\textsc{DT}_{O}$ & 0.739 & 0.734 & 0.742 & 0.740 & \textit{9} \\
     & $\textsc{kNN}_{O}$ & 0.801 & 0.795 & 0.798 & 0.801 & \textit{9} \\
    \midrule
     & \textsc{DT} & 0.770 & 0.766 & 0.772 & 0.770 & - \\
     & \textsc{kNN} & 0.809 & 0.808 & 0.811 & 0.810 & - \\
    \bottomrule
    \end{tabular}
    \label{tab:results}
\end{table}

\textbf{Quantitative Results.}
Tab.~\ref{tab:results} reports the mean predictive performance, and the number of pivots of the various predictive models\footnote{For $\textsc{DT}_{P}$ and \textsc{PTC}, we trained each best configuration with 50 different random states. Since standard deviations resulted to be negligible, we report only the average result.}.
\textsc{D2} has the highest performance, at the cost of being not interpretable.
However, a not markedly inferior performance is achieved by  \textsc{PivotTree} predictor, i.e., \textsc{PTC}, that only requires 9 pivots (6 of which are shown in Fig.~\ref{fig:pivotree_example}).
The third best performer is
\textsc{PivotTree} used as selector for a \textsc{DT}, i.e., $\textsc{DT}_{P}$.
Unfortunately, such performance is accompanied by high complexity, as $\textsc{DT}_{P}$ requires 47 pivots.
Finally, $\textsc{kNN}_{P}$, i.e.,  \textsc{PivotTree} used as selector for a \textsc{kNN} is the predictor requiring the smallest number of pivots.
Overall, \textsc{PivotTree} both employed as selector and predictor leads to competitive results compared to \textsc{D2}.
We underline how \textsc{PTC} has the best trade-off between accuracy and complexity, showing competitive results w.r.t. the fine-tuned \textsc{D2} but providing an interpretable predictor through its pivot structure, and the low number of pivots adopted. Remarkably, selecting the set of pivots $P$ through \textsc{PivotTree} leads to a  \textsc{kNN} and a \textsc{DT} which are better than those resulting using the ground-truth prototypes, especially for the \textsc{DT} case, underlying that those instances which for humans are clear examples, perhaps didactic examples, of certain cases, are not necessarily the best ones to discriminate through an automatic AI system. Finally, we remark that the performance of any \textsc{PivotTree}-based model is better than those of the \textsc{kNN} and \textsc{DT} classifiers directly trained on embeddings.

\begin{figure}[t!]
    \centering
   \includegraphics[width=\linewidth]{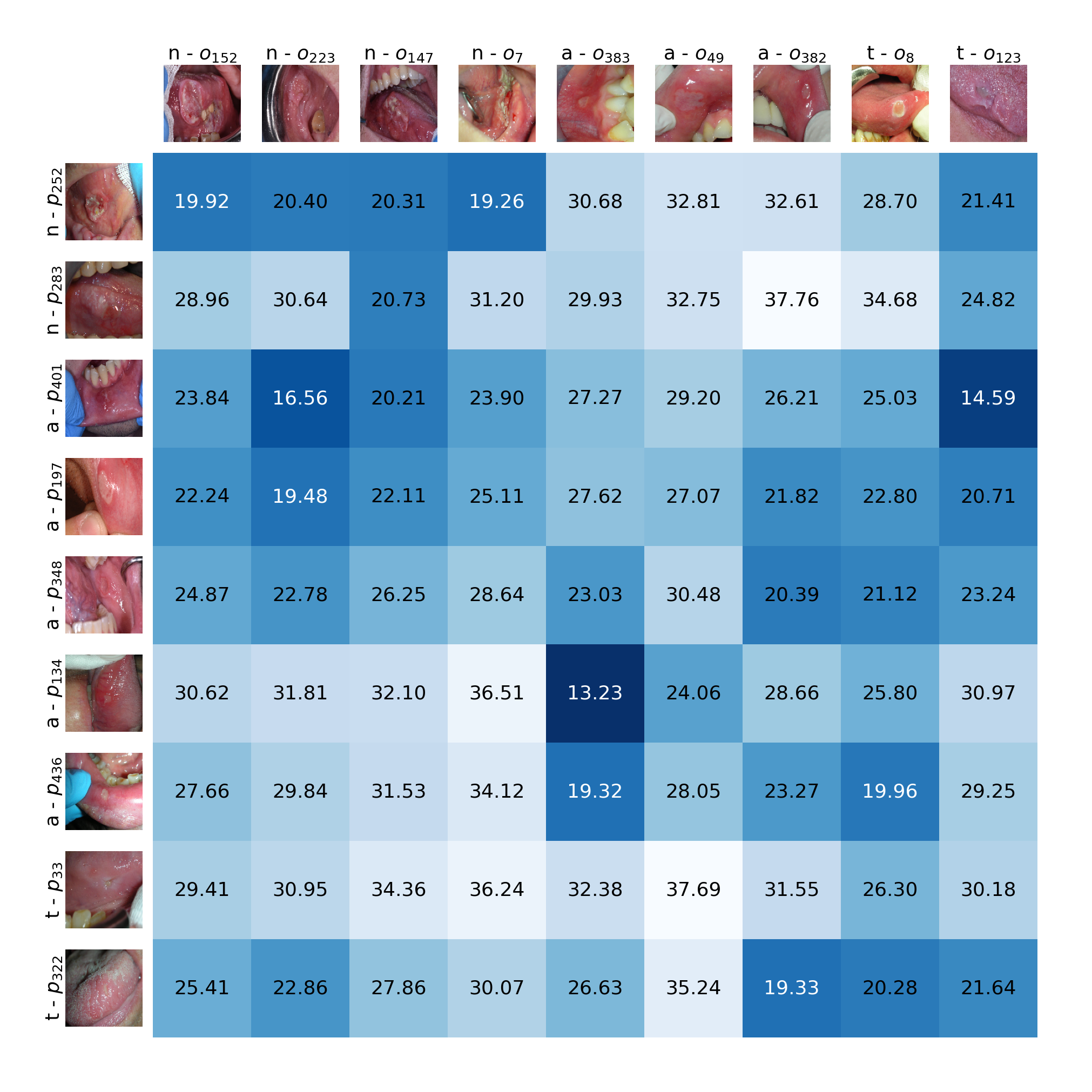}
   \caption{\textsc{PivotTree} pivots (rows) and ground-truth prototypes (columns) comparison as Euclidean distances on \textsc{D2} embedding. The darker the color the more similar are a pivot and a ground truth prototype. The first letter identifies the class of the instances: \textit{n}eoplastic, \textit{a}phthous, and \textit{t}raumatic.}
    \label{fig:heatmap_euclid}
\end{figure}

\begin{figure}[t!]
    \centering
    \includegraphics[width=\linewidth]{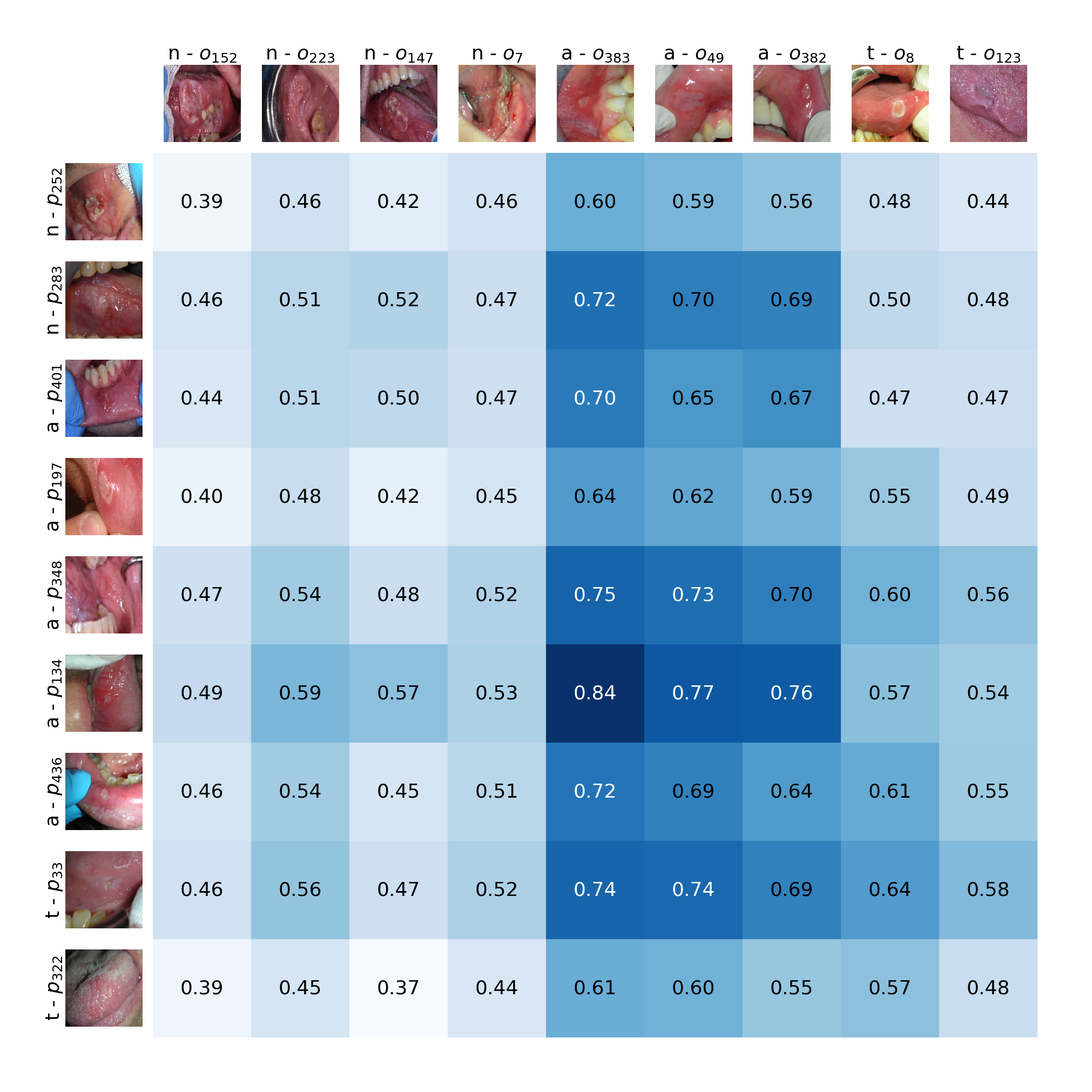}
    \caption{\textsc{PivotTree} pivots (rows) and ground-truth prototypes (columns) comparison as SSIM on raw regions of interest. Same rules from Fig.~\ref{fig:heatmap_euclid} apply.}
    \label{fig:heatmap_ssim}
\end{figure}

\textbf{Pivot-Prototypes Comparison.}
We provide here a quantitative comparison in terms of similarities between the pivots selected through \textsc{PTC} $P$ with the ground-truth prototypes $O$. 
In particular, we consider as similarity measures the Euclidean distance on the \textsc{D2} embeddings, and the Structural Similarity~\cite{structsim2004} on the original images.
For the latter, we first resize the images regions of interest to 300x300 pixels.
SSIM identifies changes in structural information by capturing the inter-dependencies among similar pixels, especially when they are spatially close. 
In Figures~\ref{fig:heatmap_euclid} and~\ref{fig:heatmap_ssim} we report two heatmaps highlighting the similarities between the \textsc{PivotTree} pivots (rows) and ground-truth prototypes (columns), on Euclidean and SSIM similarity, respectively.
Darker colors indicate higher similarity.
For the similarity comparison through Euclidean distance, we specify that the average distance between each pair of instances in the DoctOral-AI training set is $26.90 \pm 6.48$. 
When examining the average distance between pivot and ground-truth pairs w.r.t. each class in the heatmap, we find the following values: $23.93$ for \textit{neoplastic}, $24.65$ for \textit{aphthous}, and $24.60$ for \textit{traumatic}. 
This shows how the mean pairwise distances within individual classes are generally close to the overall mean pairwise distance.
Pivots and ground-truth prototypes tend to not present robust similarities. 
Furthermore, we notice how for pivots $p_{403}$ and $p_{238}$, both members of \textit{aphthous} class, the most similar ground-truth prototypes belong to a different class. 
On the other hand, for the other pivots, the closest ground-truth counterpart is consistently one of the same class, sometimes with a very high similarity: some examples are $p_{134}$ with $o_{382}$ and $p_{403}$ and $o_{223}$. A different tendency can be observed in Fig.~\ref{fig:heatmap_ssim} when using SSIM: the average SSIM w.r.t. each class is $0.46$ for \textit{neoplastic}, $0.70$ for \textit{aphthous}, and $0.57$ for \textit{traumatic}, with a mean similarity in the overall training set of $0.58 \pm 0.10$. 
This highlights a notably high internal similarity for the \textit{aphthous} class. 
As evident from Fig.~\ref{fig:heatmap_ssim}, the highest similarity is always observed when comparing pivots with the \textit{aphthous} ground-truth prototypes, differently from Fig.~\ref{fig:heatmap_euclid} which shows higher variability across classes more oriented towards the right matching.
This comparison corroborates the idea of relying on the Euclidean distance on the \textsc{D2} embedding space for \textsc{PivotTree}.

\section{Conclusion}
\label{sec:conclusion}
We have discussed \textsc{PivotTree} application in the case of oral lesion prediction, showing its superiority as a predictor w.r.t. other simple interpretable models and as selector when paired with such simple models trained on the similarity space induced by the selected pivots. 
Furthermore, we have compared expert-selected prototypes with \textsc{PTC}-selected pivots, highlighting how a strong similarity can be observed in some of the pairs. Given its flexibility, \textsc{PivotTree} lends itself to be applied for several other diagnostic task in the healthcare sector. Future investigations include testing \textsc{PivotTree} on medical data of different modalities (time-series, text reports, tabular data) in order to assess its performance, comparing it against neural prototype-based approaches for medical data as explored in~\cite{Kim2021XProtoNetDI,Singh2021genprotonet} and evaluating the interpretability of identified pivots through human subjects. Furthermore, other splitting strategies could be analyzed, one being a direct comparison between pairs of pivots as shown in \textsc{ProximityTree} models~\cite{Lucas2019proximityfor} or attempting to generate instead of select the \textsc{PivotTree}  model~\cite{guidotti2024gentree}.

\begin{credits}
\subsubsection{\ackname} 
This work has been partially supported by the European Community Horizon~2020 
programme under the funding schemes
ERC-2018-ADG G.A. 834756 ``XAI: Science and technology for the eXplanation of AI decision making'', ``INFRAIA-01-2018-2019 – Integrating Activities for Advanced Communities'', G.A. 871042, ``SoBigData++: European Integrated Infrastructure for Social Mining and Big Data Analytics'',  
by the European Commission under the NextGeneration EU programme – National Recovery and Resilience Plan (Piano Nazionale di Ripresa e Resilienza, PNRR) – Project: ``SoBigData.it – Strengthening the Italian RI for Social Mining and Big Data Analytics'' – Prot. IR0000013 – Avviso n. 3264 del 28/12/2021, and M4C2 - Investimento 1.3, Partenariato Esteso PE00000013 - ``FAIR - Future Artificial Intelligence Research'' - Spoke 1 ``Human-centered AI'',
M4 C2, Investment 1.5 "Creating and strengthening of "innovation ecosystems", building "territorial R\&D leaders", project "THE - Tuscany Health Ecosystem", Spoke 6 "Precision Medicine and Personalized Healthcare", by the Italian Project Fondo Italiano per la Scienza FIS00001966 MIMOSA, by the "Reasoning" project, PRIN 2020 LS Programme, Project number 2493 04-11-2021, by the Italian Ministry of Education and Research (MIUR) in the framework of the FoReLab project (Departments of Excellence), by the European Union, Next Generation EU, within the PRIN 2022 framework project PIANO (Personalized Interventions Against Online Toxicity) under CUP B53D23013290006.

\subsubsection{\discintname}
The authors have no competing interests to declare that are
relevant to the content of this article.
\end{credits}

\bibliographystyle{splncs04}
\bibliography{biblio_short}
 
\end{document}